\documentclass[runningheads]{llncs}
\usepackage{graphicx}

\usepackage{tikz}
\usepackage{comment} 
\usepackage{amsmath,amssymb} %
\usepackage{color}
\usepackage{diagbox}
\usepackage{microtype}
\usepackage{hyperref}
\usepackage{epsfig}
\usepackage{subfigure}
\usepackage{booktabs}
\usepackage{arydshln}
\usepackage{multirow}
\usepackage{booktabs}
\usepackage{caption}
\usepackage{float}
\usepackage{xcolor}
\usepackage{gensymb}
\usepackage[export]{adjustbox}

\usepackage[width=122mm,left=12mm,paperwidth=146mm,height=193mm,top=12mm,paperheight=217mm]{geometry}

\begin{document}
\pagestyle{headings}
\mainmatter
\def\ECCVSubNumber{2784}  %

\title{Learning Feature Descriptors using\\ Camera Pose Supervision}

\titlerunning{Learning Feature Descriptors using Camera Pose Supervision}
\author{Qianqian Wang\inst{1,2} \and
Xiaowei Zhou\inst{3} \and
Bharath Hariharan\inst{1}\and
Noah Snavely \inst{1,2} \\
\medskip
$^1$Cornell University ~~~~ $^2$Cornell Tech~~~~$^3$Zhejiang University}

\authorrunning{Q. Wang et al.}

\institute{}

\maketitle

\begin{abstract}
Recent research on learned visual descriptors has shown promising improvements in correspondence estimation, a key component of many 3D vision tasks. However, existing descriptor learning frameworks typically require ground-truth correspondences between feature points for training, which are challenging to acquire at scale. 
In this paper we propose a novel \emph{weakly-supervised} framework that can learn feature descriptors \emph{solely} from relative camera poses between images.
To do so, we devise both a new loss function that exploits the epipolar constraint given by camera poses, and a new model architecture that makes the whole pipeline differentiable and efficient.
Because we no longer need pixel-level ground-truth correspondences, our framework
opens up the possibility of training on much larger and more diverse datasets for better and unbiased descriptors.
We call the resulting descriptors \textbf{CA}mera \textbf{P}ose \textbf{S}upervised, or \textbf{CAPS}, descriptors.
Though trained with weak supervision, CAPS descriptors outperform even prior \emph{fully-supervised} descriptors and achieve state-of-the-art performance on a variety of geometric tasks.\footnote{Project page: \url{https://qianqianwang68.github.io/CAPS/}}

\end{abstract}

\section{Introduction}
Finding local feature correspondence is a fundamental component of many computer vision tasks, such as structure from motion~(SfM)~\cite{schonberger2016structure} and visual localization~\cite{sattler2018benchmarking}.
Recently, learned feature descriptors~\cite{mishchuk2017hardnet,simo2015discriminative,yi2016lift} have shown significant improvements over hand-crafted ones~\cite{bay2006surf,ke2004pca,lowe2004distinctive} on standard benchmarks. However, other recent work has observed that, when applied to real-world unseen scenarios, learned descriptors do not always generalize well~\cite{luo2018geodesc,schoenberger2017comparative}. 

One potential cause of such limited generalization is the insufficiency of high-quality training data in both quantity and diversity~\cite{schoenberger2017comparative}. 
Ideally, one would train descriptors on fully accurate, dense ground-truth correspondence between image pairs. However, it is hard to collect such data for real imagery, and only a few datasets of this form exist~\cite{chang2017matterport3d,dai2017scannet}.
As an alternative, many previous methods resort to SfM datasets that provide pseudo ground-truth correspondences given by matched and reconstructed feature points \cite{luo2018geodesc,mishchuk2017hardnet,ono2018lf,yi2016lift}, but these correspondences are sparse and potentially biased by the keypoints used in the SfM pipeline. Another option for obtaining correspondence annotations is synthetic image pairs warped by homographies~\cite{detone2018superpoint,melekhov2019dgc}. However, homographies do not capture the full range of geometric and photometric variations observed in real images. 

In this paper, we address the challenge of limited training data in descriptor learning by relaxing this requirement of ground-truth pixel-level correspondences. We propose to learn descriptors solely from relative camera poses between pairs of images. 
Camera poses can be obtained via a variety of non-vision-based sensors, such as IMUs and GPS, and can also be estimated reliably using SfM pipelines~\cite{schonberger2016structure}. By reducing the supervision requirement to camera poses, it becomes possible to learn better descriptors on much larger and more diverse datasets.

\begin{figure}[t]
    \centering
    \includegraphics[width=\linewidth]{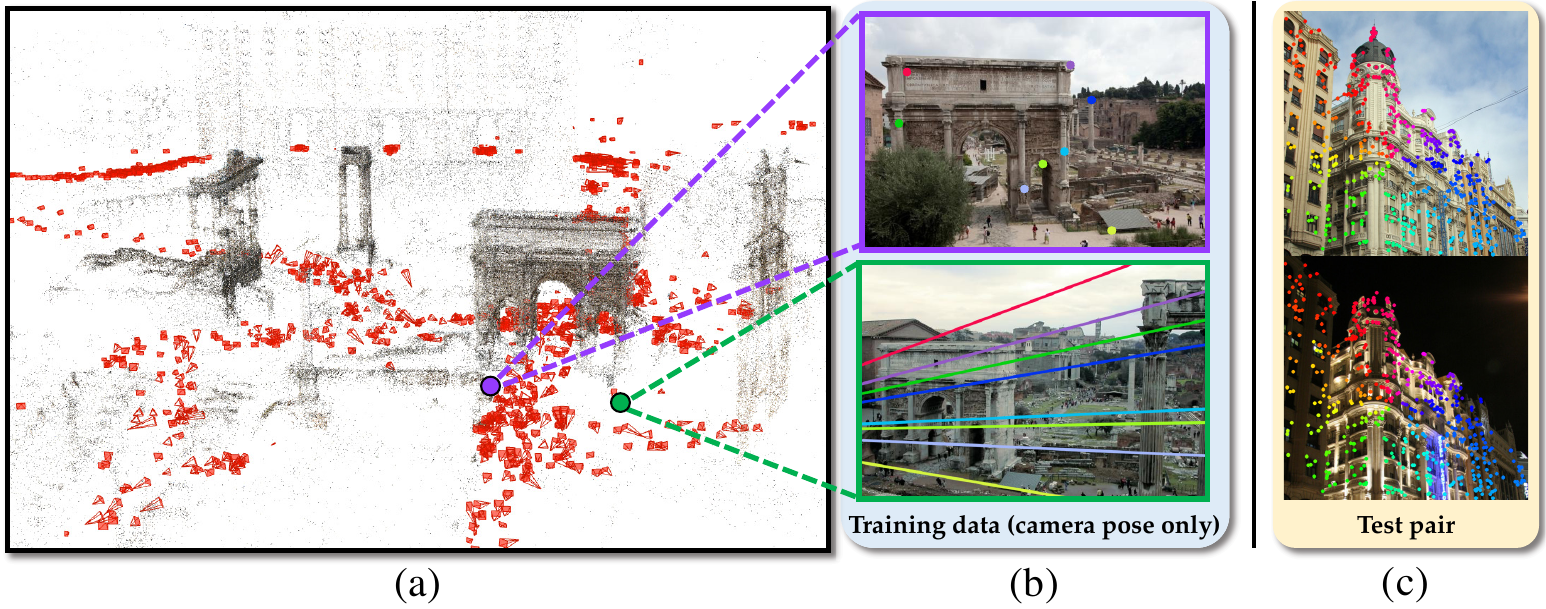}
    \caption{\small \textbf{Overview of our method}. Our model can learn descriptor using only relative camera poses (e.g., from SfM reconstructions~\textbf{(a)}). Knowing camera poses, we obtain epipolar constraints illustrated in \textbf{(b)}, where points in the first image correspond to the epipolar lines in same color in the second image.
    We utilize such epipolar constraints as our supervision signal~(see Fig.~\ref{fig:loss_function}). 
    \textbf{(c)} shows that at inference, our descriptors establish reliable correspondences even for challenging image pairs.}
    \label{fig:teaser}
\end{figure}

However, existing metric learning based methods for learning descriptors cannot utilize camera poses as supervision, as the triplet or contrastive losses used in such methods cannot be defined with respect to camera poses. Hence, we propose a novel framework to leverage camera pose supervision. Specifically, 
we translate the relative camera pose between an image pair into an epipolar constraint on pixel locations of matched points as our supervision signal~(Fig. \ref{fig:loss_function}). 
The remaining challenge is to make the locations of matched points differentiable with respect to descriptors for training, for which we introduce a new differentiable matching layer~(Fig. \ref{fig:differentiable matching}).
To further reduce the computation cost and accelerate training, we use a coarse-to-fine matching scheme~(Fig.~\ref{fig:coarse-to-fine}) that computes the correspondence at a lower resolution, then locally refines at a finer scale.

Once trained, our system can generate dense feature descriptors for an arbitrary input image, which can then be combined with existing keypoint detectors for downstream tasks.  
Despite the fact that we only train with \textit{weak} camera pose supervision, our learned descriptors are on par with or even outperform prior \emph{fully-supervised} state-of-the-art methods that train with ground-truth correspondence annotations. Furthermore, while enabling training with solely camera poses, our framework can also be trained with ground-truth correspondences, yielding even better results.

Fig.~\ref{fig:teaser} summarizes our approach. To conclude, our main contributions are:
\begin{itemize}
\item We show that camera poses alone suffice to learn good descriptors, which has not been explored in the literature to our knowledge.
\item To enable learning from camera poses, we depart from existing metric learning-based approaches and design a novel loss function as well as a new, efficient network architecture.
\item We achieve state-of-the-art performance across a range of geometric tasks.
\end{itemize}
\section{Related Work}

\noindent \textbf{Descriptor Learning}.
The dominant paradigm for learning feature descriptors is essentially deep metric learning~\cite{choy2016universal}, which encourages matching points to be close whereas non-matching points to be far away in the feature space. Various loss functions~
(e.g., pairwise and triplet loss~\cite{balntas2016learning,choy2016universal,detone2018superpoint,kumar2016learning,zhang2019learning}, structured loss~\cite{mishchuk2017hardnet,oh2016deep,sohn2016improved,Tian2017l2net})
have been developed. Based on the input type, current descriptor learning approaches roughly fall into two categories, \textit{patch-based} and \textit{dense} descriptor methods. Patch-based methods~\cite{balntas2016learning,ebel2019beyond,he2018local,keller2018learning,luo2018geodesc,mishchuk2017hardnet,mishkin2018repeatability,mukundan2019explicit,ono2018lf,simo2015discriminative,Tian2017l2net,yi2016lift} produce a feature descriptor for each \textit{patch} defined by a keypoint detector, which can be viewed as direct counterparts for hand-crafted feature descriptors~\cite{bay2006surf,calonder2010brief,lowe2004distinctive,rublee2011orb}. Dense descriptor methods~\cite{christiansen2019unsuperpoint,detone2018superpoint,dusmanu2019d2,fathy2018hierarchical,liu2019gift,r2d2,schmidt2016self} instead use fully-convolutional neural networks~\cite{long2015fully} to extract dense feature descriptors for the whole image in one forward pass. Our method gives dense descriptors, and unlike the prior work that requires ground-truth correspondence annotations to train, we are able to learn descriptors from the weak supervision of camera pose.

\medskip
\noindent \textbf{Correspondence Learning.}
Our differentiable matching layer is related to the correlation layer and cost volume that are widely used to compute stereo correspondences~\cite{chang2018pyramid,kendall2017gcnet} or optical flow~\cite{fischer2015flownet,ilg2017flownet,sun2018pwc} in a differentiable manner.
However, the search space in these problems is limited to either a single scanline or a local patch, while in wide-baseline matching we must search for matches over the whole image. This necessitates the efficient coarse-to-fine architecture we use.
Our method is also related to weakly-supervised semantic correspondence approaches~\cite{jeon2018parn,kim2018recurrent,novotny2018self,rocco2017convolutional,zhou2016learning}. However, they usually assume a simpler parametric transformation between images and tolerate much coarser correspondences than what is required for geometric tasks. Recent work~\cite{melekhov2019dgc,rocco2018neighbourhood} explores dense geometric correspondence, but focuses on global optimization of the estimated correspondences rather than the descriptors themselves.
In contrast to these prior work, we propose a new architecture that is more suitable for descriptor learning.

\medskip
\noindent 
\textbf{Epipolar Constraint}.
Epipolar constraint has been shown to be useful for learning local features~\cite{jafarian2018monet,guandao2019learning} and optical flow~\cite{zhong2019unsupervised}. MONET~\cite{jafarian2018monet} proposes the epipolar divergence for learning semantic keypoints, but this loss does not apply to dense descriptor learning. \cite{guandao2019learning} leverages epipolar constraints to generate pseudo-groundtruth correspondences but this process is non-differentiable. In contrast, we enable differentiable training of dense descriptors using the epipolar constraint.

\section{Method}
Given only image pairs with camera pose, standard deep metric learning methods do not apply.
Therefore,  we devise a new method to exploit the geometric information of camera pose for descriptor learning.
Specifically, we translate relative camera pose into an epipolar constraint between image pairs, and enforce the predicted matches to obey this constraint~(Sec.~\ref{sec: loss_formulation}). Since this constraint is imposed on pixel coordinates, we must make the coordinates of correspondences differentiable with respect to the feature descriptors. For this we devise a differentiable matching layer~(Sec.~\ref{sec: expected representation}). To further improve efficiency, we introduce a coarse-to-fine architecture~(Sec.~\ref{sec: coarse-to-fine}) to accelerate training, which also boosts the descriptor performance. We elaborate on our method below.  

\subsection{Loss Formulation}
\label{sec: loss_formulation}
\begin{figure}[t]
    \centering
    \includegraphics[width=0.5\linewidth]{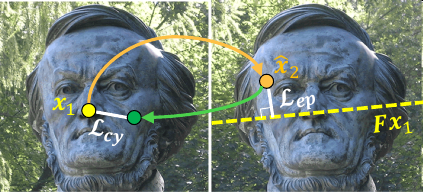}
    \caption{\small \textbf{Epipolar loss and cycle consistency loss.}
    $\mathbf{x}_1$~(yellow) is the query point, and $\mathbf{\hat{x}}_2$~(orange) is the predicted correspondence. The epipolar loss $\mathcal{L}_{\textit{ep}}$ is the distance between $\mathbf{\hat{x}}_2$ and ground-truth epipolar line $\mathbf{Fx}_1$. The cycle consistency loss $\mathcal{L}_{\textit{cy}}$ is the $L_2$ distance between $\mathbf{x}_1$ and its forward-backward corresponding point~(green).
    }
    \label{fig:loss_function}
\end{figure}

Our training data consists of image pairs with relative camera poses.
To train our correspondence system with such data, 
we propose to use two complimentary loss terms: a novel epipolar loss, and a cycle consistency loss~(Fig.~\ref{fig:loss_function}).

Given the relative pose and camera intrinsics for a pair of images $\mathbf{I}_1$ and $\mathbf{I}_2$, one can compute the fundamental matrix $\mathbf{F}$. The epipolar constraint states that $\mathbf{x}_2^{T}\mathbf{F}\mathbf{x}_1 = 0$ holds if $\mathbf{x}_1$ and $\mathbf{x}_2$ is a true match, where $\mathbf{F}\mathbf{x}_1$ can be interpreted as the epipolar line corresponding to $\mathbf{x}_1$ in $\mathbf{I}_2$.\footnote{For simplicity, we use the same symbols for homogeneous and Cartesian coordinates.}
We treat $\mathbf{x}_1$ as the query point and re-fashion this constraint into an epipolar loss based on the distance between the predicted correspondence location and the ground-truth epipolar line:
\begin{equation}
    \mathcal{L}_{\textit{ep}}(\mathbf{x}_1) = \text{dist}(h_{1\rightarrow 2}(\mathbf{x}_{1}), \mathbf{F}\mathbf{x}_1),
\end{equation}
where $h_{1\rightarrow 2}(\mathbf{x}_1)$ is the predicted correspondence in $\mathbf{I}_2$ for the point $\mathbf{x}_1$ in $\mathbf{I}_1$, and $\text{dist}(\cdot, \cdot)$ is the distance between a point and a line.

The epipolar loss alone only encourages a predicted match to lie on the epipolar line, rather than near the ground-truth correspondence location (which is at an unknown location on the line). To provide additional supervision, we additionally introduce a cycle consistency loss. 
This loss encourages the forward-backward mapping of a point to be spatially close to itself~\cite{wang2019learning}:
\begin{equation}
    \mathcal{L}_\textit{cy}(\mathbf{x}_1) = ||h_{2\rightarrow 1}(h_{1\rightarrow 2}(\mathbf{x}_{1})) -  \mathbf{x}_1||_2.
\end{equation}
This term encourages the network to find true correspondences and suppress other outputs, especially those that satisfy the epipolar constraint alone.

\medskip
\noindent \textbf{Full Training Objective.}
For each image pair, our total objective is a weighted sum of epipolar and cycle consistency losses, totaled over $n$ sampled query points:
\begin{equation}
    \mathcal{L}(\mathbf{I}_1, \mathbf{I}_2) = \sum_{i=1}^n [\mathcal{L}_{\textit{ep}}(\mathbf{x}_{1}^i) + \lambda \mathcal{L}_{\textit{cy}}(\mathbf{x}_{1}^i)], 
    \label{eq:loss_both}
\end{equation}
where $\mathbf{x}_{1}^i$ is the $i$-th training point in $\mathbf{I}_1$, and $\lambda$ is a weight for the cycle consistency loss term. 
At the end of Sec.~\ref{sec: expected representation}, we further show how we can reweight individual training instances in Eq.~(\ref{eq:loss_both}) %
to improve training.

\subsection{Differentiable Matching Layer}
\label{sec: expected representation}
The objective defined above is a simple function of the pixel locations of the predicted correspondences. 
Minimizing this objective through gradient descent therefore requires these locations to be differentiable with respect to the network parameters.
Many prior methods establish correspondence by identifying nearest neighbor matches, which unfortunately is a non-differentiable operation.

\begin{figure}[t]
    \centering
    \subfigure[differential matching layer]{
    \includegraphics[width=0.45\linewidth,valign=b]{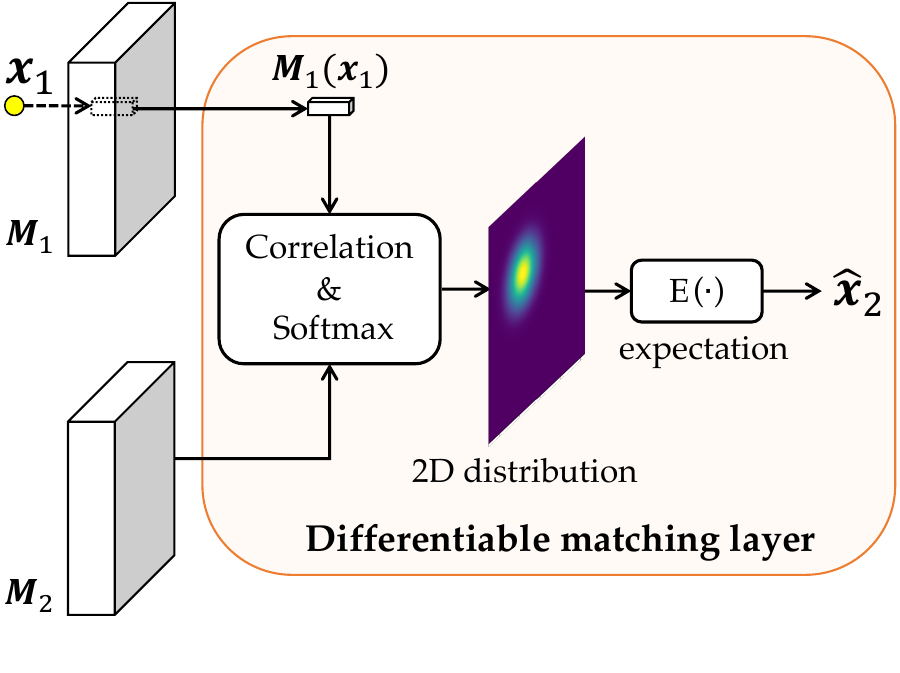}
    \label{fig:differentiable matching}}
    \qquad
    \subfigure[coarse-to-fine architecture]{
    \includegraphics[width=0.43\linewidth,valign=b]{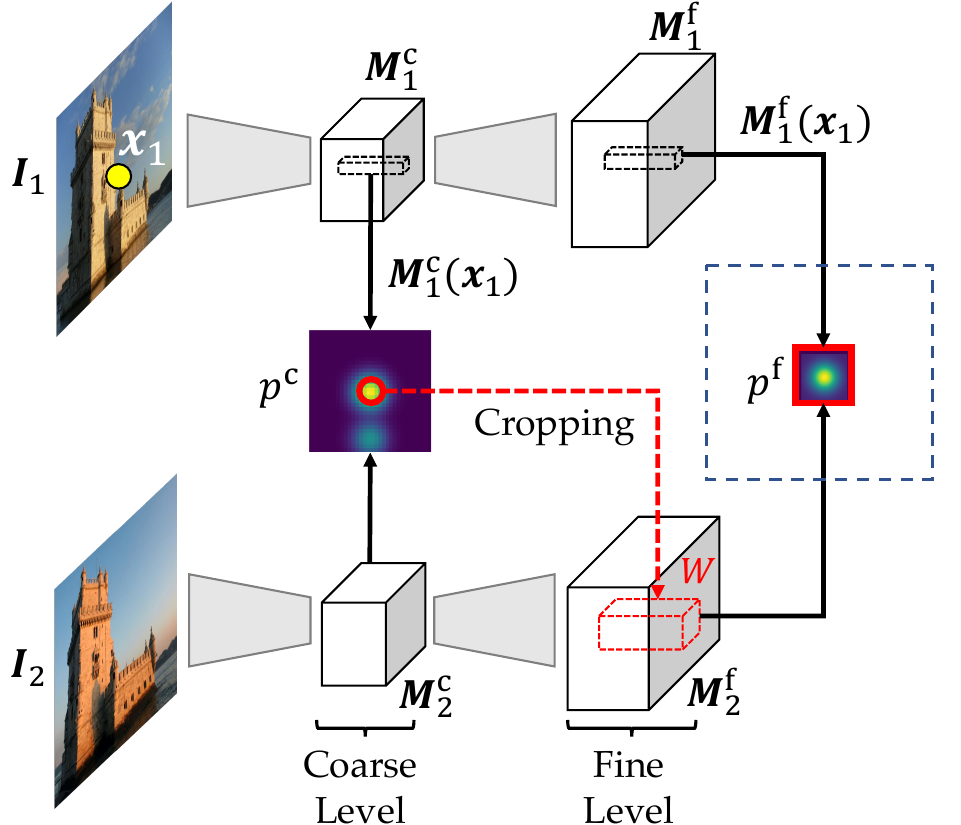}
    \label{fig:coarse-to-fine}}
    
    \caption{\small \textbf{Network architecture design.} \textbf{(a) differentiable matching layer}. 
    For a query point, its correspondence location is represented as the expectation of a distribution computed from the correlation between feature descriptors.
    \textbf{(b) The coarse-to-fine module}. We use the location of highest probability at coarse level~(red circle) to determine the location of a local window $W$ at the fine level. During training, we compute the correspondence locations at both coarse and fine level from distribution $p^c$ and $p^f$, respectively, and impose our loss functions on both. This allows us to train both coarse- and fine-level features simultaneously.
    }
    \label{fig:architecture}
\end{figure}

To address this challenge,  we propose a differentiable matching layer, illustrated in Fig.~\ref{fig:differentiable matching}.
Given a pair of images, we first use convolutional networks with shared weights to extract dense feature descriptors $\mathbf{M}_1$ and $\mathbf{M}_2$.
To compute the correspondence for a query point $\mathbf{x}_1$ in $\mathbf{I}_1$, we correlate the feature descriptor at $\mathbf{x}_1$, denoted by $\mathbf{M}_1(\mathbf{x}_1)$, with all of $\mathbf{M}_2$. Following a 2D softmax operation~\cite{Goodfellow-et-al-2016}, we obtain a distribution over 2D pixel locations of $\mathbf{I}_2$, indicating the probability of each location being the correspondence of $\mathbf{x}_1$. We denote this probability distribution as $p(\mathbf{x}|\mathbf{x}_1,\mathbf{M}_1, \mathbf{M}_2)$:
\begin{equation}
    p(\mathbf{x}|\mathbf{x}_1, \mathbf{M}_1, \mathbf{M}_2) = \frac{\exp{(\mathbf{M}_1(\mathbf{x}_1)^\text{T} \mathbf{M}_2(\mathbf{x}))}}{\sum_{\mathbf{y}\in \mathbf{I}_2}\exp{(\mathbf{M}_1(\mathbf{x}_1)^\text{T}\mathbf{M}_2(\mathbf{y}))}},
    \label{eq:prob}
\end{equation}
where $\mathbf{y}$ varies over the pixel grid of $\mathbf{I}_2$.
A single 2D match can then be computed as the expectation of this distribution:
\begin{equation}
     \hat{\mathbf{x}}_2= h_{1\rightarrow 2}(\mathbf{x}_1) = \sum_{\mathbf{x} \in \mathbf{I}_2} \mathbf{x} \cdot p(\mathbf{x}|\mathbf{x}_1, \mathbf{M}_1, \mathbf{M}_2) .
\end{equation}
This makes the entire system end-to-end trainable. Since the correspondence location is computed from the correlation between feature descriptors, enforcing it to be correct would facilitate descriptor learning.

\medskip
\noindent \textbf{Leveraging Uncertainty during Training}.
This differentiable matching also provides an interpretable measure of \emph{uncertainty}.
For each query point $\mathbf{x}_1$, we can calculate the total variance $\sigma^2(\mathbf{x}_1)$ as an uncertainty measure, which is defined as the trace of the covariance matrix of the 2D distribution $p(\mathbf{x} | \mathbf{x}_1, \mathbf{M}_1, \mathbf{M}_2)$. 
High variance indicates multiple or diffuse modes, signifying an unreliable prediction.

This uncertainty can help identify unreliable correspondences and improve training.
In particular, due to the lack of ground-truth correspondence annotations, it is unknown if a query point has a true correspondence in the other image during training~(which could be missing due to occlusion or truncation). Minimizing the loss for such points can lead to incorrect training signals. To alleviate this issue, we reweight the losses for each individual point using the total variance defined above, resulting in the final weighted loss function:
\begin{equation}
    \mathcal{L}(\mathbf{I}_1, \mathbf{I}_2) = \sum_{i=1}^n \frac{1}{\sigma(\mathbf{x}_{1}^i)}[\mathcal{L}_{\textit{ep}}(\mathbf{x}_{1}^i) + \lambda \mathcal{L}_{\textit{cy}}(\mathbf{x}_{1}^i)],
    \label{eq:final_loss}
\end{equation}
where the weight $1 / \sigma(\mathbf{x}_{1}^i)$ are normalized so that they sum up to one. 
This weighting strategy weakens the effect of infeasible and non-discriminative training points, which we find to be critical for rapid convergence. Prior work~\cite{jeon2019joint,novotny2018self} on semantic correspondence leverages the uncertainty in a similar way, but their uncertainty is predicted using extra network parameters whereas ours is directly derived from the learned descriptors.

\subsection{Coarse-to-Fine Architecture}
\label{sec: coarse-to-fine}
During training, we impose supervision only on sparsely sampled query points for each pair of images. While the computational cost is made manageable in this way, having to search correspondence over the entire image space is still costly. To overcome this issue, we propose a coarse-to-fine architecture that significantly improves computational efficiency, 
while preserving the resolution of learned descriptors. Fig.~\ref{fig:coarse-to-fine} illustrates the coarse-to-fine module. Instead of generating a flat feature descriptor map, we produce both coarse-level feature descriptors $\mathbf{M}_1^{\text{c}}, \mathbf{M}_2^{\text{c}}$ and  fine-level feature descriptors $\mathbf{M}_1^{\text{f}}, \mathbf{M}_2^{\text{f}}$.

Coarse-to-fine matching works as follows. Given a query point $\mathbf{x}_1$, we first compute the distribution $p^{\text{c}}(\mathbf{x}|\mathbf{x}_1, \mathbf{M}_1^{\text{c}}, \mathbf{M}_2^{\text{c}})$ over \textit{all locations} of the coarse feature map. At the fine level, on the contrary, we compute the fine-level distribution only in a \textit{local window} $W$ centered at the highest probability location in the coarse-level distribution (with coordinates rescaled appropriately).
Given coarse- and fine-level distributions, correspondences at both levels can be computed. 
We then impose our loss function~(Eq.~(\ref{eq:final_loss})) on correspondences at both levels, which allows us to train both coarse and fine features descriptors simultaneously.

This architecture allows us to learn high-resolution descriptors without evaluating full correlation between large feature maps, significantly reducing computational cost. In addition, as observed by Liu et al. \cite{liu2008sift}, we find that coarse-to-fine reasoning not only improves efficiency but also boosts matching accuracy~(Sec.~\ref{sec:ablation}). By concatenating both coarse- and fine-level descriptors, we obtain the final hierarchical descriptors~\cite{fathy2018hierarchical} that capture both abstract and detailed information.

\subsection{Discussion}
\label{sec: method-discussion}
\noindent \textbf{Effectiveness of Epipolar Constraint}. 
The seemingly weak epipolar constraint actually provides empirically sufficient supervision for descriptor learning, as suggested by results in Sec.~\ref{sec:experiments}.  
One key reason is that the epipolar constraint suppresses a large number of incorrect correspondence---i.e., every point not on the epipolar line.
Moreover, among all valid predictions that satisfy the epipolar constraint, true correspondences are most likely to have similar feature encodings given their local appearance similarity.
Therefore, by aggregating such a geometric constraint over all training data, the network learns to encode the similarity between true correspondences, leading to effective learned descriptors.

\medskip
\noindent \textbf{Training with Ground-truth Correspondence Annotations}. Although the focus of this paper is on learning from camera poses alone, our system can also be trained with ground-truth correspondence annotations when such data is available. In this case, we can replace our loss functions with an $L_2$ distance between the pixel locations of the predicted and ground-truth correspondence. As shown in Fig.~\ref{fig:ablation}, our method trained with groundtruth correspondences achieves even better performance than our method trained with camera poses, with both outperforming prior fully supervised methods.

\medskip
\noindent \textbf{Matching at Test Time}.
The descriptors learned by our system can be integrated in standard feature matching pipelines. Given a detected keypoint, feature vectors in the coarse and fine feature maps are extracted by interpolation and concatenated to form the final descriptor. We then match features using the standard Euclidean distance between them.

\subsection{Implementation Details}
\label{sec:implementation details}
\noindent \textbf{Architecture}. 
We use a ImageNet-pretrained ResNet-50~\cite{deng2009imagenet,he2016deep,paszke2017automatic} architecture, truncated after \texttt{layer3}, as our backbone.
With an additional convolutional layer we obtain the coarse-level feature map. The fine-level feature map is obtained by further convolutional layers along with up-sampling and skip-connections.
The sizes of the coarse- and fine-level feature map are $1/16$ and $1/4$ of the original image size, respectively. They both have a feature dimensionality of $128$. The size of the local window $W$ at fine level is $1/8\times$ the size of the fine-level feature map. 

\medskip
\noindent \textbf{Training Data}.
We train using the MegaDepth dataset~\cite{li2018megadepth}, which consists of 196 different scenes reconstructed from over 1M internet photos using COLMAP~\cite{schonberger2016structure}.
130 out of 196 scenes are used for training and the rest are for validation and testing.
This gives us millions of training pairs with known camera poses. We train our system on these pairs using only the provided camera poses and intrinsics.

\medskip
\noindent \textbf{Training Details}.
We train the network using Adam~\cite{kingma2014adam} with a base learning rate of $10^{-4}$.
The weight $\lambda$ for the cycle consistency term is set to $0.1$. $n=500$ query points are used in each training image pair due to memory constraints. These query points consist of 90\% SIFT~\cite{lowe2004distinctive} keypoints and 10\% random points. 

\medskip
\noindent For more implementation details, please refer to the supplementary material.

\section{Experimental Results}\label{sec:experiments}
To evaluate our descriptors, referred to as CAPS, we conduct three sets of experiments:
\begin{enumerate}
\item \textbf{Feature matching experiments:} 
The most direct evaluation of CAPS is in terms of how accurately they can be matched between images. 
We evaluate both sparse and dense feature matching on the HPatches dataset~\cite{hpatches_2017_cvpr}.
\item \textbf{Experiments on downstream tasks:} Feature matches are rarely the end-goal. 
Instead, they form a core part of many 3D reconstruction tasks. 
We evaluate the impact of CAPS on downstream tasks
(homography estimation on HPatches as well as relative pose estimation on MegaDepth~\cite{li2018megadepth} and ScanNet~\cite{dai2017scannet}) 
and 3D reconstruction
(as part of an SfM pipeline in the ETH local feature benchmark~\cite{schoenberger2017comparative}).
\item \textbf{Ablation study:} We evaluate the impact of each proposed contribution using the HPatches dataset.
\end{enumerate}

\subsection{Feature Matching Results}
\label{sec:image matching}
We evaluate our descriptors on both sparse and dense feature matching on the HPatches dataset~\cite{hpatches_2017_cvpr}.
HPatches is a homography dataset containing 116 sequences, where 57 sequences have illumination changes and 59 have viewpoint changes. 

\medskip
\noindent\textbf{Sparse Feature Matching}. Given a pair of images, we extract keypoints in both images and match them using feature descriptors. We follow the same evaluation protocol as in D2-Net~\cite{dusmanu2019d2} and use the mean matching accuracy (MMA) as the evaluation metric. The MMA score is defined as the average percentage of correct matches per image pair under a certain pixel error threshold. Only mutual nearest neighbor matches are considered. 

We combine CAPS with SIFT~\cite{lowe2004distinctive} and SuperPoint~\cite{detone2018superpoint} keypoints which are representative of hand-crafted and learned keypoints, respectively.
We compare to several baselines: Hessian
affine detector~\cite{mikolajczyk2004scale} with RootSIFT descriptor~\cite{lowe2004distinctive,arandjelovic2012three}~(HesAff + RootSIFT), 
HesAffNet~\cite{mishkin2018repeatability} regions with HardNet++ descriptors~\cite{mishchuk2017hardnet}~(HAN + HN++),
DELF~\cite{delf}, SuperPoint~\cite{detone2018superpoint}, LF-Net~\cite{ono2018lf}, multi-scale
D2-Net~\cite{dusmanu2019d2}~(D2-Net MS), SIFT detector with ContextDesc descriptors~\cite{luo2019contextdesc}~(SIFT + ContextDesc), as well as R2D2~\cite{r2d2}.
\begin{figure}[t]
\centering
\begin{minipage}{0.73\linewidth}
\centering
\includegraphics[width=\linewidth]{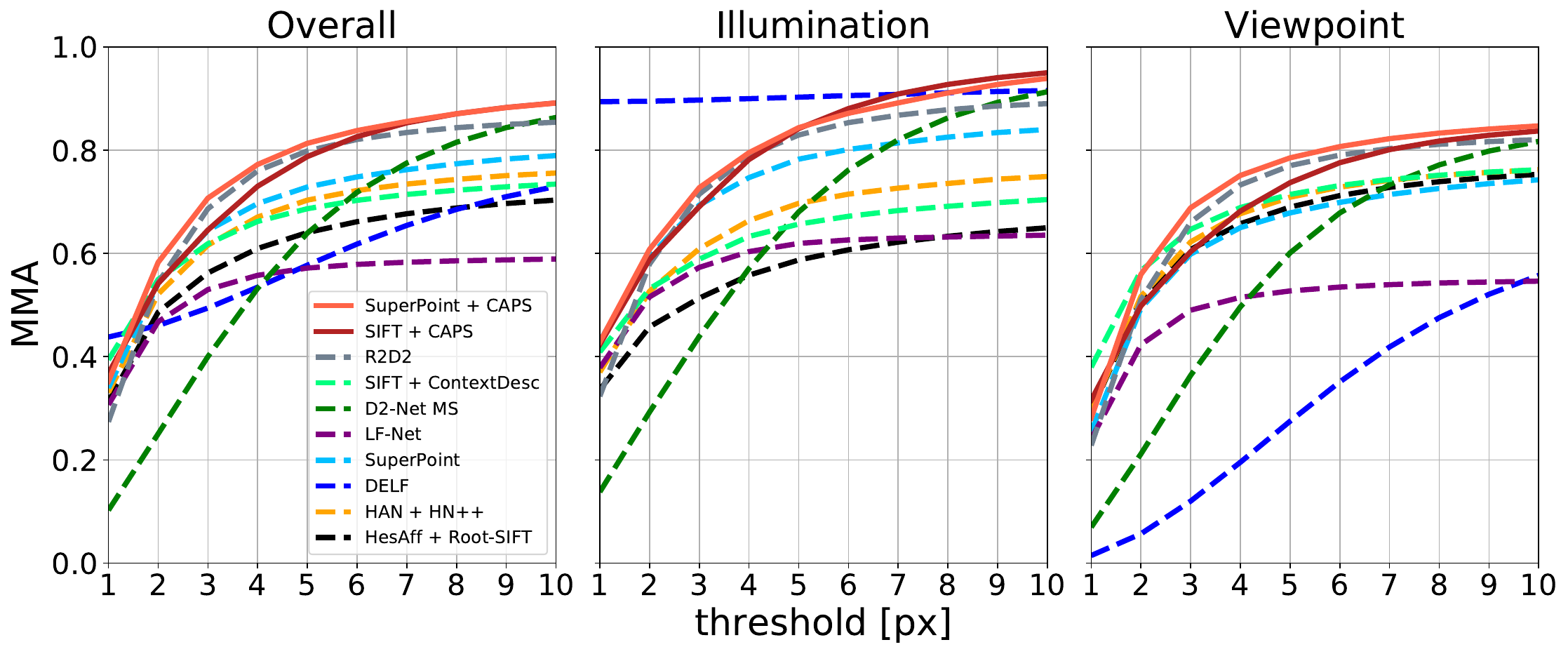}
\end{minipage}
\begin{minipage}{0.26\textwidth}
\centering
    \centering
    \small
    \resizebox{1\textwidth}{!}{%
    \begin{tabular}{lrr}
    \toprule
    \textbf{Methods} & \# \textbf{Feature} & \# \textbf{Match}\\
    \midrule
     HesAff + Root-SIFT & 6.7K & 2.8K\\
    HAN + HN++ & 3.9K & 2.0K \\
    LF-Net & 0.5K & 0.2K\\
    SuperPoint & 1.7K & 0.9K\\
    DELF & 4.6K & 1.9K\\
    SIFT + ContextDesc & 4.4K & 1.9K\\
    D2-Net MS & 8.3K & 2.8K \\
    R2D2 & 5.0K & 1.8K\\
    \hline
    SuperPoint + CAPS & 1.7K & 0.9K \\
    SIFT + CAPS & 4.4K & 1.5K\\
    \bottomrule
    \end{tabular}
    }
\end{minipage}
\caption{\small
\textbf{Mean matching accuracy~(MMA) on HPatches~\cite{hpatches_2017_cvpr}}. For each method, we show the MMA with varying pixel error thresholds. We also report the mean number of detected features and mutual nearest neighbor matches. 
With SuperPoint~\cite{detone2018superpoint} keypoints, our approach achieves the best overall performance after $2$px.}
\label{fig:mma_hpatches}
\end{figure}

Fig.~\ref{fig:mma_hpatches} shows MMA results on the HPatches dataset. We report results for the whole dataset, as well as for subsets corresponding to illumination and viewpoint changes. Following D2-Net \cite{dusmanu2019d2}, we additionally present the mean number of detected features per image and mutual nearest neighbor matches per pair. With SuperPoint keypoints CAPS achieves the best overall performance, and with SIFT keypoints CAPS also achieves competitive performance. In addition, with the same detectors, CAPS shows clear improvements over its counterparts~(``SIFT + CAPS'' vs.\ ``SIFT + ContextDesc'', ``SuperPoint + CAPS'' vs.\ ``SuperPoint'').

\medskip
\noindent\textbf{Dense Feature Matching}. To evaluate our dense matching capability, we extract keypoints on image grids in the first image and find their nearest neighbor match in the full second image. The percentage of correct keypoints (PCK) metric~\cite{choy2016universal,long2014convnets,zhou2015flowweb} is used to measure performance: the predicted match for a query point is deemed correct if it is within a certain pixel threshold of the true match.

\begin{figure}[t]
    \centering
    \subfigure[PCK comparison]{
    \includegraphics[width=0.28\linewidth,valign=c]{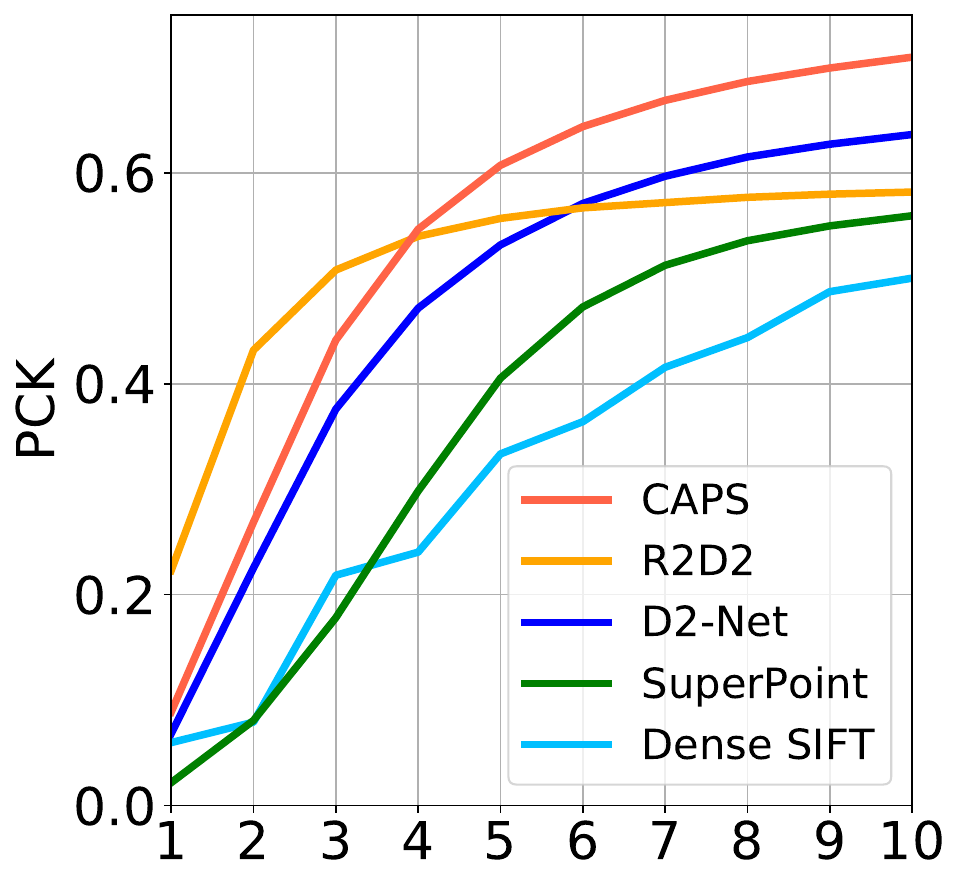}\label{fig:pck}}
    \subfigure[Qualitative result]{
    \includegraphics[width=0.68\linewidth,valign=c]{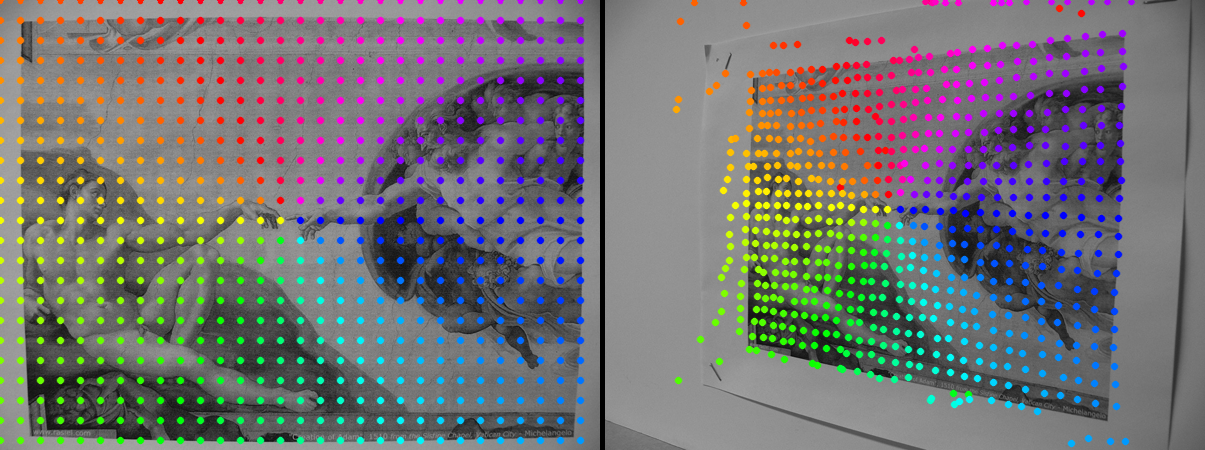}
    \label{fig:dense correspondence}}
    \caption{\small \textbf{Dense feature matching on HPatches.} \textbf{(a)} PCK comparison. CAPS outperforms other methods at larger pixel thresholds~($>$ 4px). \textbf{(b)} Qualitative result of dense feature matching.
    Color indicates correspondence.}
    \label{fig:dense feature matching}
\end{figure}

We compare to baseline methods that produce dense descriptors: Dense SIFT~\cite{lowe2004distinctive}, SuperPoint~\cite{detone2018superpoint}, D2-Net~\cite{dusmanu2019d2} and R2D2~\cite{r2d2}. Fig.~\ref{fig:pck} shows the mean PCK over all image pairs on HPatches. 
CAPS achieves the overall best performance and is only worse than R2D2~\cite{r2d2} at small thresholds ($\leq$ 4px). This is because the R2D2 we use here computes descriptor maps at the full input image resolution, whereas ours are 4x downsampled. Fig.~\ref{fig:dense correspondence} shows the qualitative performance of our dense correspondence.

\subsection{Results on Downstream Tasks}
Next, we evaluate how well CAPS facilitates downstream tasks. We focus on two tasks related to two-view geometry estimation: homography estimation and relative camera pose estimation, and a third task related to 3D reconstruction.

\medskip
\noindent \textbf{Homography Estimation.}
We use the same HPatches dataset as in Sec.~\ref{sec:image matching} for the homography estimation task. We follow the corner correctness metric used in SuperPoint~\cite{detone2016deephomo,detone2018superpoint}. The four corners of one image are transformed to the other image using the estimated homography and compared with the corners computed using the groundtruth homography. 
The estimated homography is deemed correct if the average error of the four corners is less than $\epsilon$ pixels. 

Following SuperPoint~\cite{detone2018superpoint}, we extract a maximum of 1,000 keypoints from each image, and robustly estimate the homography from mutual nearest neighbor matches. 
The comparison of homography accuracy between CAPS and other methods is shown in Tab.~\ref{tab:homography_estimation}. 
As can be seen, CAPS improves over both SIFT and SuperPoint descriptors. With SuperPoint keypoints, CAPS achieves the overall best performance even without training on annotated correspondences.
\begin{table}[h]
    \begin{center}
    \footnotesize
    \setlength{\tabcolsep}{5pt}
    \caption{\small \textbf{Homography estimation accuracy~[\%] at 1, 3, 5 pixels on HPatches.} CAPS with SuperPoint keypoints  achieves the overall best performance.
    }
    \label{tab:homography_estimation}
    \resizebox{0.55\linewidth}{!}{%
    \begin{tabular}{lccc}
    \toprule
    \textbf{Methods}         & $\epsilon=1$  & $\epsilon=3$  & $\epsilon=5$ \\
    \midrule
    SIFT~\cite{lowe2004distinctive}  & 40.5         & 68.1         & 77.6 \\
    LF-Net~\cite{ono2018lf}          & 34.8         & 62.9         & 73.8 \\
    SuperPoint~\cite{detone2018superpoint}  & 37.4         & 73.1         & 82.8 \\
    D2-Net~\cite{dusmanu2019d2}     & 16.7         & 61.0         & 75.9 \\
    ContextDesc~\cite{luo2019contextdesc} & 41.0         & 73.1         & 82.2 \\
    R2D2~\cite{r2d2}                    & 40.0          & \textbf{75.0}          & 84.7 \\
    \midrule
    CAPS w/ SIFT kp.     & 34.6         & 72.2         & 81.7 \\
    CAPS w/ SuperPoint kp. & \textbf{44.8} & 74.5 & \textbf{85.7} \\
    \bottomrule
    \end{tabular}}
    \end{center}
    \vspace{-1em}
\end{table}

\begin{table}[t]
    \begin{center}
    \footnotesize
    \setlength{\tabcolsep}{3.3pt}
    \caption{\small \textbf{Relative pose estimation accuracy on ScanNet~\cite{dai2017scannet} and MegaDepth~\cite{li2018megadepth}.} Each cell shows the accuracy of estimated rotations and translations (as \textit{rotation accuracy / translation accuracy}). Each value shown is the percentage of pairs with relative pose error under a certain threshold ($5^\circ$ for ScanNet and $10^\circ$ for MegaDepth). \textbf{Higher is better.}  $d_{\text{frame}}$ represents the interval between frames. Larger frame intervals imply harder pairs for matching. 
}
     \label{tab:relative_pose}
    \resizebox{\linewidth}{!}{%
    \begin{tabular}{lcccccc}
    \toprule
    \multirow{2}*{
    \textbf{Methods}
    } &\multicolumn{3}{c}{\textbf{Accuracy on ScanNet [\%]}} & \multicolumn{3}{c}{\textbf{Accuracy on MegaDepth [\%]}}\\ %
    \cmidrule(l){2-4} \cmidrule(l){5-7}
    & \textit{$d_{\text{frame}}$ = 10} & \textit{$d_{\text{frame}}$ = 30} & \textit{$d_{\text{frame}}$ = 60} & \textit{easy} & \textit{moderate} & \textit{hard} \\ %
    \midrule
    SIFT~\cite{lowe2004distinctive}& 
    91.0 / 14.1 & 65.1 / 15.6 & 41.4 / 11.9&
    58.9 / 20.2 & 26.9 / 11.8 & 13.6 / 9.6\\
    SIFT w/ ratio test~\cite{lowe2004distinctive}& 
    91.2 / 15.9 & 67.1 / 19.8 & 44.3 / 15.9&
    63.9 / 25.6 & 36.5 / 17.0 & 20.8 / 13.2\\
    SuperPoint~\cite{detone2018superpoint}& 
    94.4 / 17.5 & 75.9 / 26.3 & 53.4 / 22.1&
    67.2 / 27.1 & 38.7 / 18.8 & 24.5 / 14.1\\
    HardNet~\cite{mishchuk2017hardnet} & 
    95.8 / 18.2 & 79.0 / 24.7 & 55.6 / 21.8 &
    66.3 / 26.7 & 39.3 / 18.8 & 22.5 / 12.3 \\
    LF-Net~\cite{ono2018lf}&
    93.6 / 17.4 & 76.0 / 22.4 & 49.9 / 18.0 &
    52.3 / 18.6 & 25.5 / 13.2 & 15.4 / 11.1 \\
    D2-Net~\cite{dusmanu2019d2} & 
    91.6 / 13.3 & 68.4 / 19.5 & 42.0 / 14.6&
    61.8 / 23.6 & 35.2 / 19.2 & 19.1 / 12.2 \\
    ContextDesc~\cite{luo2019contextdesc} &
    91.5 / 16.3 & 73.8 / 21.8 & 51.4 / 18.5&
    68.9 / 27.1 & 43.1 / 21.5 & 27.5 / 14.1 \\
    R2D2~\cite{r2d2} & 
    \textbf{97.4} / \textbf{22.3} & \textbf{86.1} / \textbf{31.7} & \textbf{62.9} / \textbf{28.8} &
    69.4 / 30.3 & 48.3 / 23.9 & 32.6 / 17.4 \\
    \midrule
    CAPS w/ SIFT kp. &
    92.3 / 16.3 & 74.8 / 22.5 & 50.8 / 20.9 &
    70.0 / 30.5 & 50.2 / 24.8 & 36.8 / 16.1\\
    CAPS w/ SuperPoint kp. &
    96.1 / 17.1 & 79.5 / 27.2 & 59.3 / 26.1 &
    \textbf{72.9} / \textbf{30.5} & \textbf{53.5} / \textbf{27.9} & \textbf{38.1} / \textbf{19.2} \\
    \bottomrule
    \end{tabular}}
    \end{center}
    \vspace{-1em}
\end{table}

\medskip
\noindent \textbf{Relative Pose Estimation.} We also evaluate the performance of CAPS on the task of relative camera pose estimation. Note that we train only on MegaDepth~\cite{li2018megadepth} but test on both MegaDepth and ScanNet~\cite{dai2017scannet}, an indoor dataset that we use to test the generalization of CAPS.
For MegaDepth, we generate overlapping image pairs from test scenes, and
sort them into three subsets according to relative rotation angle: \textit{easy} ($[0^\circ, 15^\circ]$), \textit{moderate} ($[15^\circ, 30^\circ]$) and \textit{hard} ($[30^\circ, 60^\circ]$). 
For ScanNet, 
we follow LF-Net~\cite{ono2018lf} and randomly sample image pairs at three different frame intervals, 10, 30, and 60.
Each subset in MegaDepth and ScanNet consists of 1,000 image pairs.

To estimate relative pose, 
we first estimate the essential matrix from mutual nearest neighbor matches~(RANSAC~\cite{fischler1981ransac} is applied), and then decompose it to get the relative pose.
For SIFT~\cite{lowe2004distinctive} we additionally prune matches using the ratio test~\cite{lowe2004distinctive}, since that is the common practice for camera pose estimation (i.e, we report results of both plain SIFT and SIFT with a carefully-tuned ratio test).

\begin{figure*}[t]
    \centering
    \includegraphics[width=\linewidth]{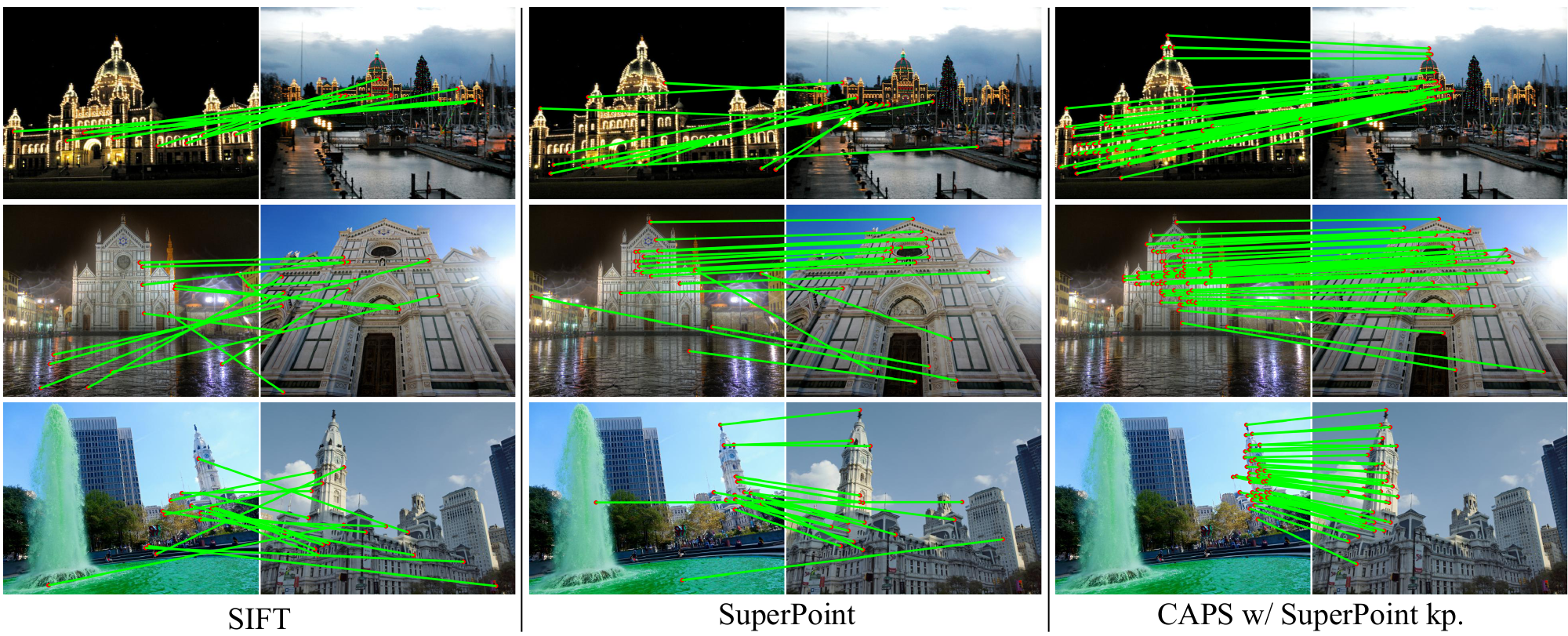}
    \caption{\small \textbf{Sparse feature matching results after RANSAC}. The test image pairs are from MegaDepth~\cite{li2018megadepth}. Green lines indicate correspondences. Our method works well even under challenging illumination and viewpoint changes.}
    \label{fig:qualitative_match}
    \vspace{-1em}
\end{figure*}

Following UCN~\cite{choy2016universal}, we evaluate the estimated camera pose using angular deviation for both rotation and translation.
We consider a rotation or translation to be correct if the angular deviation is less than a threshold, and report the average accuracy for that threshold. We set a threshold of $5^\circ$ for ScanNet and $10^\circ$ for MegaDepth, as MegaDepth is harder due to larger illumination changes. Results for all methods are reported in Tab. \ref{tab:relative_pose}. CAPS improves performance over SIFT and SuperPoint descriptors, and ``CAPS w/ SuperPoint keypoints'' outperforms all other methods but is outperformed by R2D2~\cite{r2d2} on ScanNet.
Qualitative results on MegaDepth test images are shown in Fig.~\ref{fig:qualitative_match}.

\begin{table}[t]
    \caption{\small \textbf{Evaluation on the ETH local features benchmark~\cite{schoenberger2017comparative}.} Note that SIFT and D2-Net~\cite{dusmanu2019d2} apply ratio test but other methods do not. Overall, CAPS performs on par with state-of-the-art local features on this task.}
    \centering
    \setlength{\tabcolsep}{2pt}
    \resizebox{\linewidth}{!}{%
    \begin{tabular}{llccccc}
    \toprule
     &  & \textbf{\#Registered} \ & \textbf{\#Sparse points} \ & \textbf{\#Obs.} \
         &  \textbf{Track Len.} \  & \textbf{Reproj. Err.}\ \\
    \midrule

    \textbf{Madrid} & SIFT~\cite{lowe2004distinctive}  & 500 & 116K & 734K & 6.32 & \textbf{0.61}px\\
    \textbf{Metropolis} & GeoDesc~\cite{luo2018geodesc} & 809 & 307K & 1,200K & 3.91 & 0.66px \\ 
    1,344 images & D2-Net~\cite{dusmanu2019d2} & 501 & 84K & - & \textbf{6.33} & 1.28px \\ 
    & SOSNet~\cite{tian2019sosnet} & 844 & \textbf{335K} & 1,411K & 4.21 & 0.70px\\
    & CAPS & \textbf{851} & 242K & \textbf{1,489K} & 6.16 & 1.03px\\    
    \midrule
    \textbf{Gendarmen-} & SIFT  & 1,035 & 339K & 1,872K & \textbf{5.52} & \textbf{0.70}px\\
    \textbf{markt} & GeoDesc & 1,208 & 780K & 2,903K & 3.72 & 0.74px \\ 
    1,463 images & D2-Net & 1,053 & 250K & - & 5.08 & 1.19px \\
    & SOSNet & \textbf{1,201} & \textbf{816K} & 3,255K & 3.98 & 0.77px \\
    & CAPS & 1,179 & 627K & \textbf{3,330K} & 5.31 & 1.00px\\ 
    \midrule
    \textbf{Tower of} & SIFT  & 804 & 240K & 1,863K & \textbf{7.77} & \textbf{0.62}px  \\
    \textbf{London} &  GeoDesc & 1,081 & \textbf{622K} & \textbf{2,852K} & 4.58 & 0.69px \\
    1,576 images &  D2-Net & 785 & 180K & - & 5.32 & 1.24px \\
    & CAPS  & \textbf{1,104} & 452K & 2,627K & 5.81 & 0.98px \\
    \bottomrule
    \end{tabular}}
    \label{tab:ETH_local_feature_benckmark}
    \vspace{-1em}
\end{table}

\medskip
\noindent\textbf{3D Reconstruction}.
Finally, we evaluate the effectiveness of CAPS descriptors in the context of 3D reconstruction using the ETH local features benchmark~\cite{schoenberger2017comparative}. We extract CAPS descriptors at keypoint locations provided by~\cite{schoenberger2017comparative} and feed them into the protocol. 
Following \cite{luo2018geodesc}, we do not conduct the ratio test, in order to investigate the direct matching performance of the descriptors. To quantify the quality of SfM, we report the number of registered images~(\# Registered), sparse 3D points (\#Sparse Points) and image observations (\# Obs), the mean track lengths~(Track Len.), and the mean reprojection error~(Reproj. Err.).

We use SIFT~\cite{lowe2004distinctive}, GeoDesc~\cite{luo2018geodesc}, D2-Net~\cite{dusmanu2019d2} and SOSNet~\cite{tian2019sosnet} as baselines and show the results in Tab.~\ref{tab:ETH_local_feature_benckmark}. CAPS is comparable to or even outperforms our baselines in terms of the completeness of the sparse reconstruction~(i.e., the number of registered images, sparse points and observations). However, we do not achieve the lowest reprojection error. A similar situation is observed in~\cite{luo2018geodesc,tian2019sosnet}, which can be explained by the trade-off between completeness of reconstruction and low reprojection error: fewer matches tend to lead to lower reprojection error. Taking all metrics into consideration, the performance of CAPS for SfM is competitive, indicating the advantages of CAPS even trained with only weak pose supervision.

\subsection{Ablation Analysis}
\label{sec:ablation}
In this section, we conduct ablation analysis to demonstrate the effectiveness of our proposed camera pose supervision and architectural designs. We follow the evaluation protocol in Sec.~\ref{sec:image matching} and report MMA and PCK score over all image pairs in the HPatches dataset~\cite{hpatches_2017_cvpr}. For sparse feature matching, we combine our descriptors with SIFT~\cite{lowe2004distinctive} keypoints. 
The variants of our default method~(\textit{Ours}) are introduced below. For fair comparison, we train each variant on the same training data~($\sim$20K image pairs) from MegaDepth~\cite{li2018megadepth} for 10 epochs.

\begin{figure}[t]
    \centering
    \includegraphics[width=0.72\linewidth]{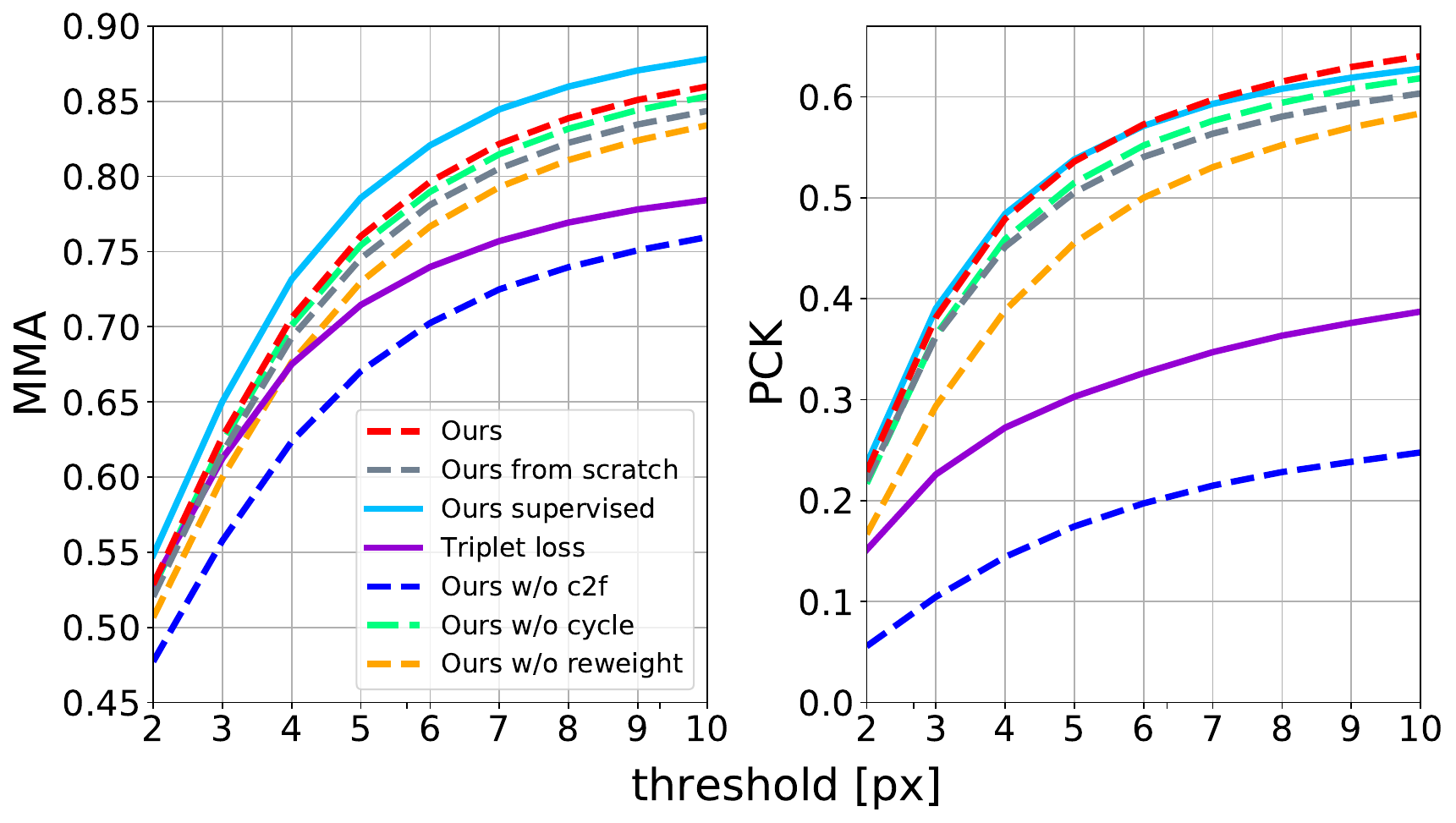}
    \caption{\small \textbf{Ablation study on HPatches}. Solid lines indicate methods trained with ground-truth correspondence; dashed lines indicate ones trained with only camera pose.}
    \label{fig:ablation}
\end{figure}

\medskip
\noindent \textbf{Variants}. \textit{Ours from scratch} is trained from scratch instead of using ImageNet~\cite{deng2009imagenet} pretrained weights. \textit{Ours supervised} is trained on sparse ground-truth correspondences provided by the SfM models of MegaDepth~\cite{li2018megadepth}. We simply change the epipolar loss to a $L_2$ loss between predicted and groundtruth correspondence locations. \textit{Triplet Loss} is also trained on sparse ground-truth correspondences, but using a standard triplet loss and a hard negative mining strategy~\cite{choy2016universal}. \textit{Ours w/o c2f} is a single-scale version of our method, where the coarse-level feature maps are removed and only the fine-level feature maps are trained and used as descriptors. \textit{Ours w/o cycle} does not use the cycle consistency loss term~($\lambda=0$), and \textit{Ours w/o reweighting} does not use the uncertainty re-weighting strategy, but uses uniform weights during training. Below we provide a detailed analysis based on these variants. The results are shown in Fig.~\ref{fig:ablation}. 

\medskip
\noindent \textbf{Analysis of Supervision Signal}. 
Both \textit{Ours supervised} and \textit{Ours} outperform the plain version of \textit{Triplet Loss}, where \textit{Ours supervised} and \textit{Triplet Loss} share the same correspondence annotations but \textit{Ours} uses only camera pose. 
\textit{Ours supervised} outperforms \textit{Triplet Loss} because of the geometric distance-based losses (as opposed to metric learning) and the coarse-to-fine architecture. Compared to \textit{Ours supervised}, the gains of \textit{Ours} decrease a bit, but our epipolar loss still leverages the rich information in the epipolar constraint and allows us to outperform \textit{Triplet Loss} and other past fully supervised work in Sec.~\ref{sec:experiments}.
In terms of loss functions, cycle consistency only provides marginal improvement, and training with only cycle consistency loss fails. This validates the importance of epipolar constraint. \textit{Ours from scratch} shows that even with randomly initialized weights, our network still succeeds to converge and learn descriptors, further validating the effectiveness of our loss functions. 

\medskip
\noindent \textbf{Analysis of Architecture Design}.  As shown in Fig.~\ref{fig:ablation}, the coarse-to-fine module significantly improves performance. Two explanations for this improvement include: 1) At the fine level,  correspondence is computed within a local window, which may reduce issues arising from multi-modal distributions compared to a flat model that computes expectations over the whole image; and 2) The coarse-to-fine module produces hierarchical feature descriptors that capture both global and local information, which may be beneficial for feature matching.

\section{Conclusion}
In this paper, we propose a novel descriptor learning framework that can be trained using only camera pose supervision. We present both new loss functions that exploit the epipolar constraints, and a new efficient architectural design that enables learning by making the correspondence differentiable. Experiments showed that our method achieves state-of-the-art performance across a range of geometric tasks, outperforming fully supervised counterparts without using any correspondence annotations for training.
In future work, we will study how to further improve invariance of the learned descriptors to large geometric transformations. It is also worth investigating if the pose supervision and traditional metric learning losses are complementary to each other, and if their combination can lead to even better performance.

\medskip
\noindent \textbf{Acknowledgements.} 
We thank Kai Zhang, Zixin Luo, Zhengqi Li for helpful discussion and comments.
This work was partly supported by a DARPA LwLL grant, and in part by the generosity of Eric and Wendy Schmidt by recommendation of the Schmidt Futures program.

\bibliographystyle{splncs04}
\bibliography{ref}

\end{document}